\documentclass[letterpaper]{article}
\usepackage{aaai20}
\usepackage{times}
\usepackage{helvet}
\usepackage{courier}
\usepackage[hyphens]{url}
\usepackage{graphicx}
\usepackage{graphicx}
\usepackage{subcaption} 
\title{Using Deep Convolutional Neural Networks to  \linebreak Detect Rendered Glitches in Video Games}

\author{
Carlos García Ling \\ carlosgl@kth.se \\ KTH Royal Institute of Technology \\ Stockholm, Sweden \And  Konrad Tollmar, Linus Gisslén  \\
\{ktollmar,lgisslen\}@ea.com\\ SEED - Electronic Arts (EA) \\ Stockholm, Sweden \\
}

\begin{document}
\maketitle
\begin{abstract}
In this paper, we present a method using Deep Convolutional Neural Networks (DCNNs) to detect common glitches in video games.  The problem setting consists of an image (800x800 RGB) as input to be classified into one of five defined classes, normal image, or one of four different kinds of glitches (stretched, low resolution, missing and placeholder textures). Using a supervised approach, we train a ShuffleNetV2 using generated data. This work focuses on detecting texture graphical anomalies achieving arguably good performance with an accuracy of 86.8\%, detecting 88\% of the glitches with a false positive rate of 8.7\%, and with the models being able to generalize and detect glitches even in unseen objects. We apply a confidence measure as well to tackle the issue with false positives as well as an effective way of aggregating images to achieve better detection in production. The main use of this work is the partial automatization of graphical testing in the final stages of video game development.

\end{abstract}

\section{Introduction and Background}

Developing video games involves many steps, starting from the concept, to the final release. Often there are hundreds of developers and artists involved when creating a modern game. In this complex process plenty of bugs can be introduced, and many of them having an negative effect on the rendered images. We refer these graphical bugs as {\it glitches} in this paper. There are several stages where graphical glitches can occur: when updating the asset database (e.g. resulting in missing textures), updating the code base (e.g. resulting in textures being corrupted), updating (graphical) drivers, cross-platform development, etc. The main method to find these visual bug is by testing the game and its input/output. Since graphics are one of the main components of a video game, it is of high importance to assure the absence of glitches or malfunctions that otherwise may reduce the player's experience.

Graphical errors are hard to programmatically detect and occur in relative small proportion. Often, they are identified by testing the game manually by a human and when detected, they are diagnosed and addressed. As they occur relative seldom it is a time consuming and mundane task. One example of graphics testing is the so called {\it smoke test}. Here a video is produced within the game by a camera traversing the environment recording different assets in the game. A human tester can then view the video and check that everything looks like expected. Although valuable, this process can be extremely time consuming and costly due to the scarcity of the glitches, and, because the lack of a systematic procedure, many of them might be overlooked. Automated testing on rendered images is crucial for an effective development process but at the same time also difficult to achieve with traditional methods.

Anomaly detection methods have proven useful when detecting rare instances \cite{chandola2009anomaly} in fields like finance \cite{ahmed2016survey}, fault detection, surveillance \cite{sultani2018real}, or medical research \cite{schlegl2017unsupervised}. With the increasing amounts of data, and computational resources more complex methods allow to detect anomalous instances even in complex data \cite{chalapathy2019deep} like image \cite{schlegl2017unsupervised} and video \cite{sultani2018real}, using unsupervised and semi-supervised approaches \cite{an2015variational,zenati2018efficient}. When looking into the problem formulated in video games, we have identified two crucial characteristics:

\begin{itemize}
    \item \textbf{Availability of data}: Video game images are generated by a rendering engine, in contrast with natural images they do not have to be sampled and potentially a big data set can be generated.
    \item	\textbf{Generation of negative samples}: When an element malfunctions in the video game and a glitch is generated, there is often a reasonable way of tracing the source. This allows the generation and reproduction of similar artifacts synthetically.
\end{itemize}

These characteristics mean that in practice both positive and negative glitch examples can be generated at a large scale for training our model. Thus, allowing for a supervised approach, meaning that we will train our model using labeled samples which usually translates on a higher performance compared to an unsupervised approach using unlabeled data. However, with this approach we will primarily be able to detect the same type of glitches as the ones used for training, meaning that other graphical glitches will remain undetected by the model. We argue that this is not necessarily a big problem as we can solve this by having an iterative process when new kinds of glitches appear they can readily be added to the training data by generating more negative examples.

Our method uses one image as an input that then will be classified as {\it correct} or {\it faulty} with 4 different kinds of glitches all related to texture malfunctions. To classify the image a DCNN was used. In recent years, since the widely used AlexNet was presented \cite{krizhevsky2012imagenet}, advances like residual networks \cite{he2016deep} or dense connections \cite{huang2017densely} have increased the performance of these methods substantially. Although these methods can be highly costly computationally, which was addressed by presenting more lightweight networks like ShuffleNet \cite{zhang2018shufflenet}.

In this work, we generate synthetic data and use it to train different DCNNs into detecting graphical malfunctions. The main contribution of this paper is showing that DCNNs can be used to detect glitches effectively, providing one step further on automating video game testing. For a more in-depth discussion, image samples, and analysis see \cite{garcia2020graphical}.
In comparison with previous work, \cite{nantes2008framework} and \cite{nantes2013neural}, where other methods such as multi-layer perceptions, self organizing maps, or corner detection are used, we propose a versatile approach that does not need any pre-processing nor extra information from the game state, but simply a game capture, and that can be configured for detecting new classes of glitches by providing adequate samples.

\section{Data}
The data used for training the models was generated using the video game engine Unity3D. Accounting for which are the most common glitches when developing video games, the data was centered on rendered texture glitches. A texture is the digital representation of the surface of an object in 3D graphics thus a fundamental part when rendering a video game scene. They define how the assets in a video game look like and when missing or incorrectly processed, they can impact the player’s video game experience negatively. Four different kinds of glitches were synthesized, that can be grouped in corrupted textures, and missing textures. Below, a description of the glitches and how they are generated:

\begin{figure}[ht]
    \centering
    \begin{subfigure}[b]{0.23\textwidth}
        \centering
        \includegraphics[width=\textwidth]{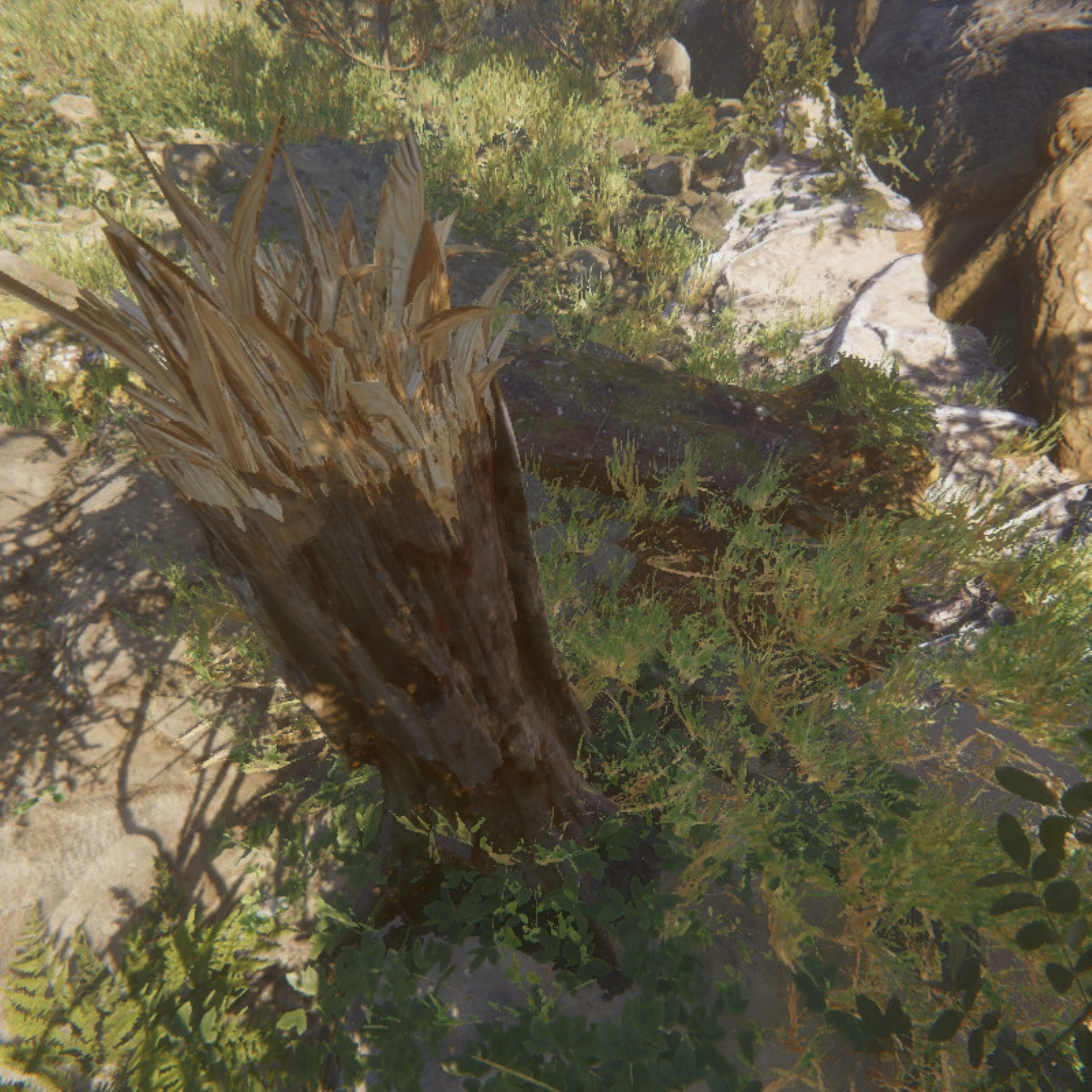}
        \caption{Normal}    
        \label{fig:Sample_Normal}
    \end{subfigure}
    \hfill
    \begin{subfigure}[b]{0.23\textwidth}  
        \centering 
        \includegraphics[width=\textwidth]{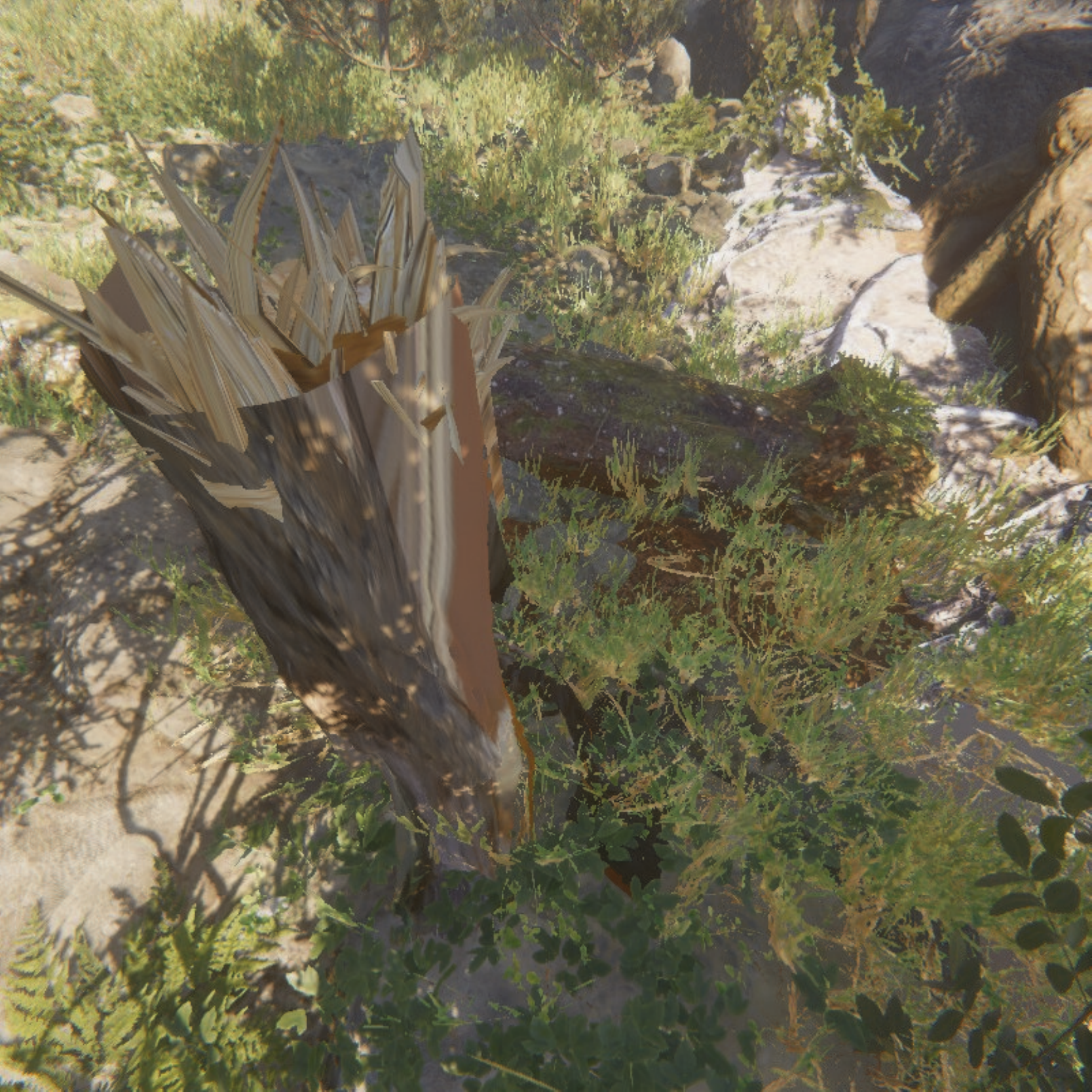}
        \caption{Stretched}    
        \label{fig:Sample_Stretched}
    \end{subfigure}
    \vskip\baselineskip
    \begin{subfigure}[b]{0.23\textwidth}
        \centering
        \includegraphics[width=\textwidth]{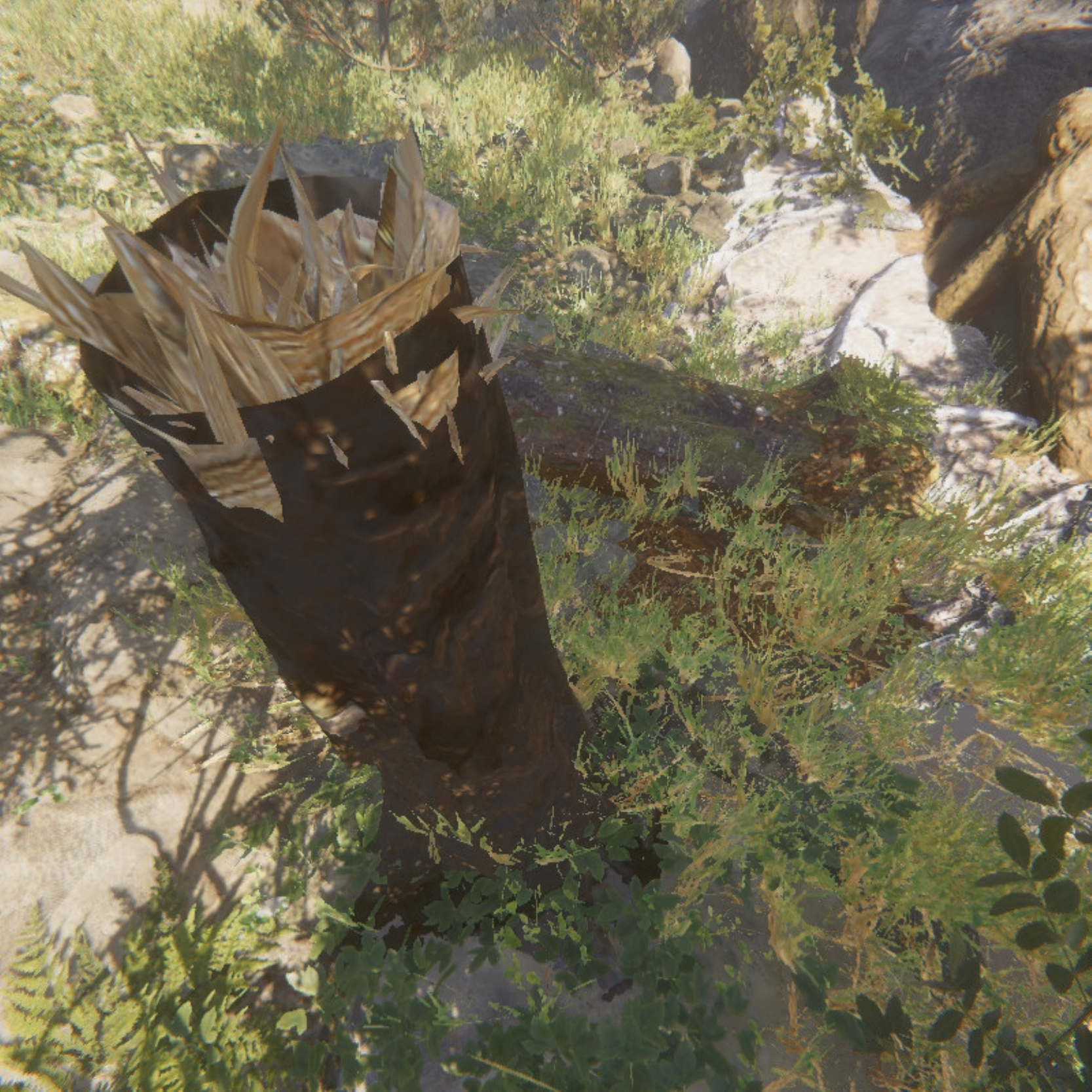}
        \caption{Low Resolution}    
        \label{fig:Sample_LowRes}
    \end{subfigure}
    \hfill
    \begin{subfigure}[b]{0.23\textwidth}  
        \centering 
        \includegraphics[width=\textwidth]{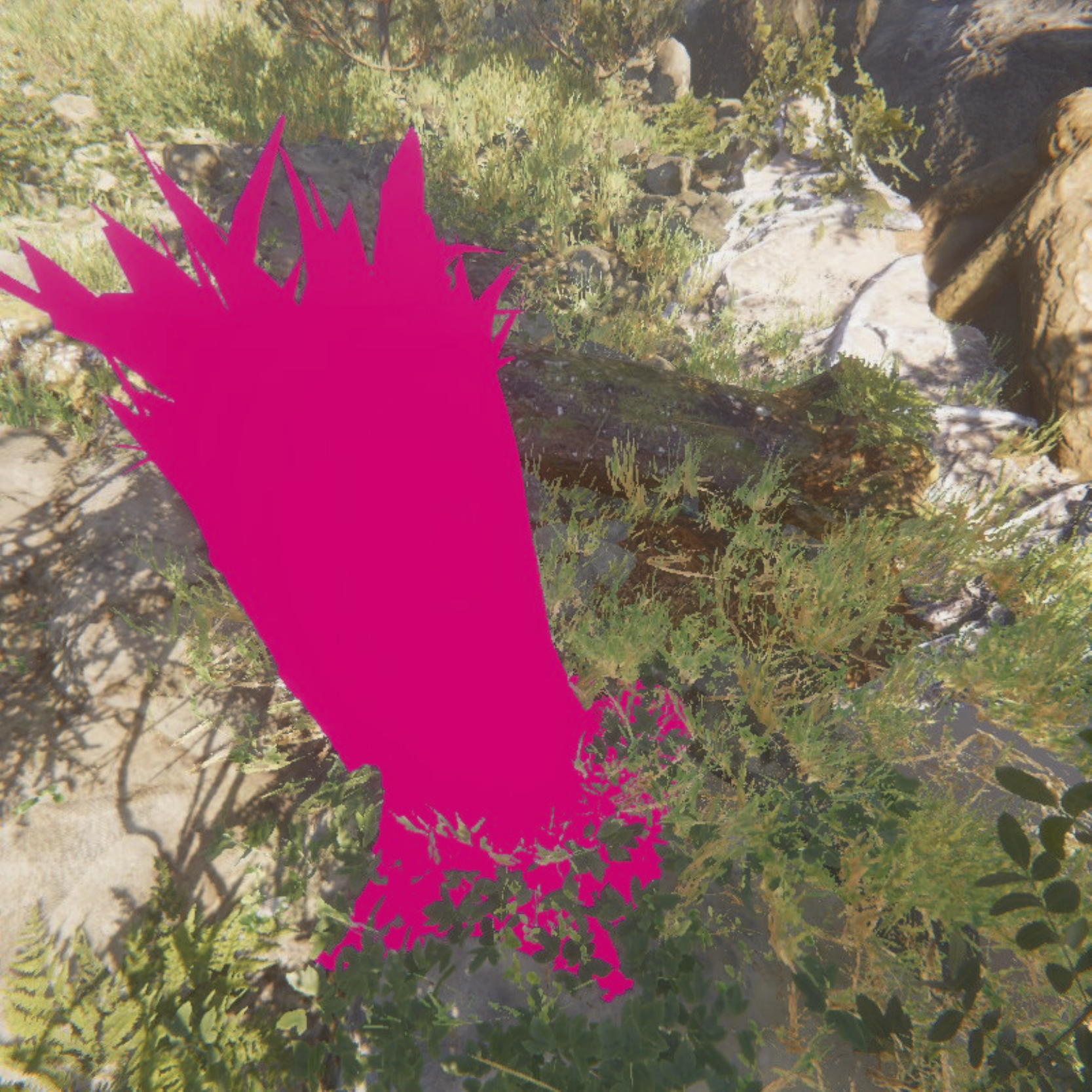}
        \caption{Missing}    
        \label{fig:Sample_Missing}
    \end{subfigure}
    \vskip\baselineskip
    \begin{subfigure}[b]{0.23\textwidth}
        \centering
        \includegraphics[width=\textwidth]{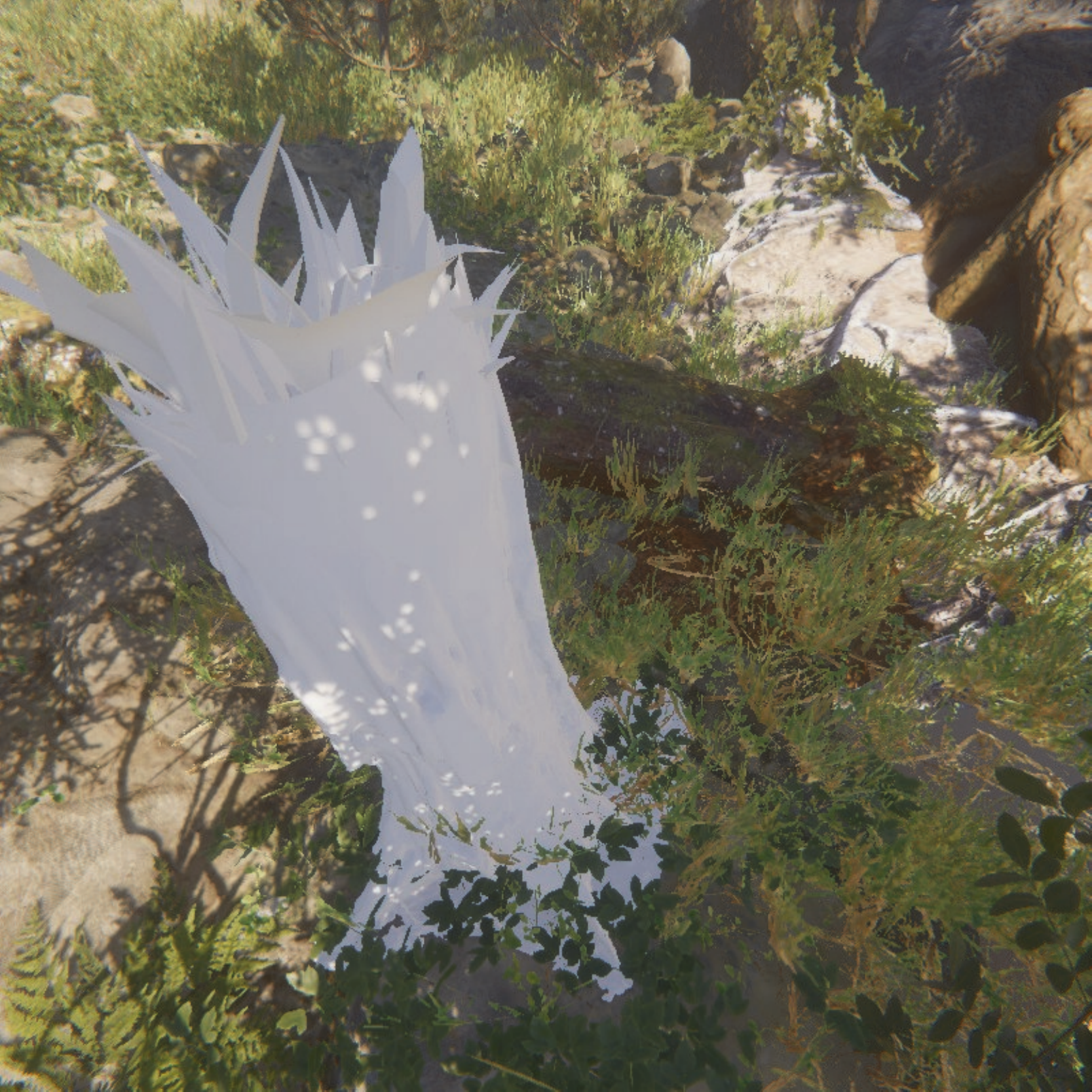}
        \caption{White Placeholder}    
        \label{fig:Sample_PlaceholderWhite}
    \end{subfigure}
    \hfill
    \begin{subfigure}[b]{0.23\textwidth}  
        \centering 
        \includegraphics[width=\textwidth]{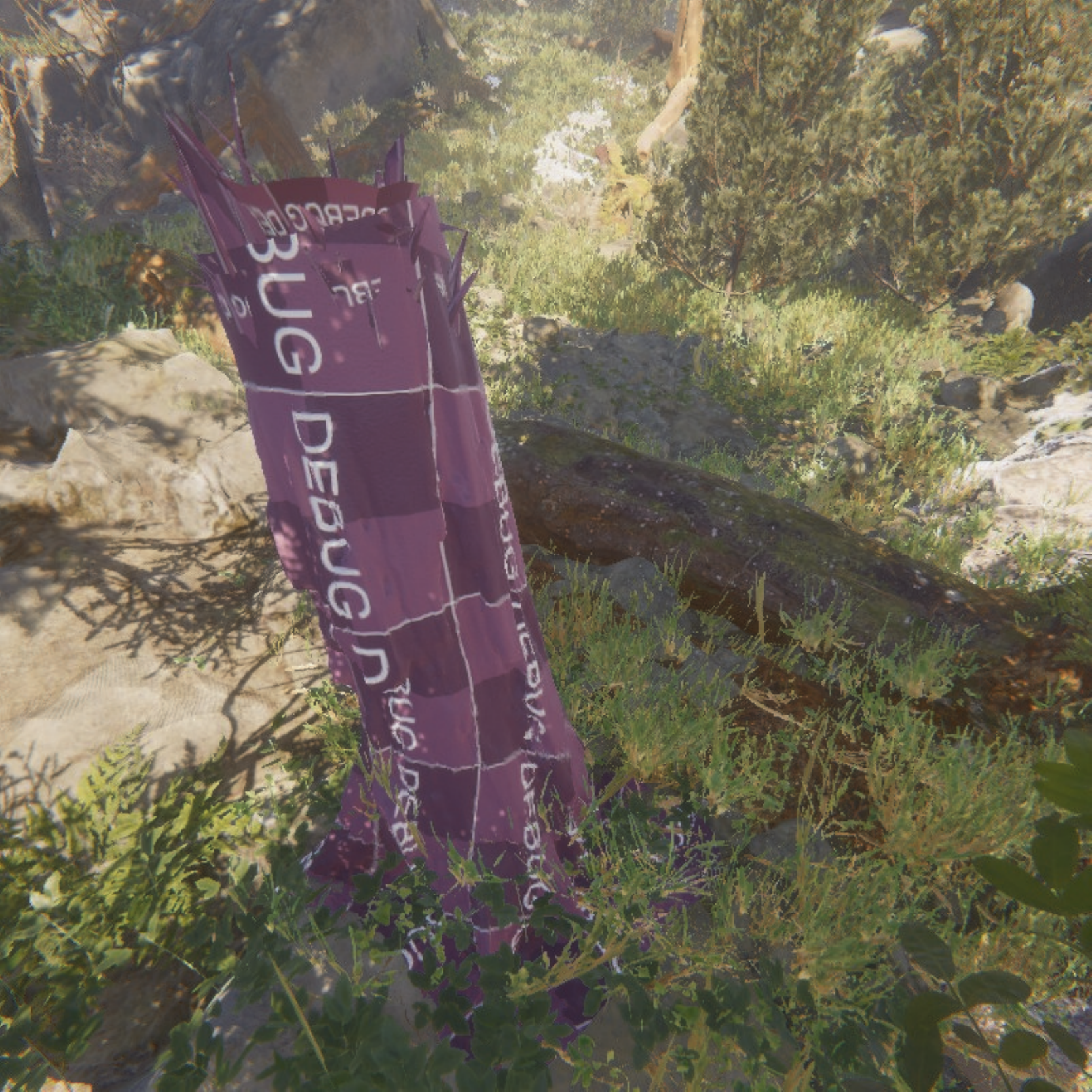}
        \caption{Pattern Placeholder}    
        \label{fig:Sample_PlaceholderPattern}
    \end{subfigure}
    \caption{Glitch samples}
    \label{fig:Samples}
\end{figure}

\textbf{Corrupted textures}: textures are still rendered in the object, but due to some error, they are {\it corrupted}. We divide them into two sub-categories:

\begin{itemize}
    \item \textbf{Stretched}: Due to errors like a wrong texture mapping to the object surface or a deformation in the object, the texture is deformed, and characteristic line patterns show in the rendered image. In Unity, the glitches were generated by scaling the texture in one random direction (see Figure \ref{fig:Sample_Stretched}). Note that in practice this is not exactly what happens in every case since the shape of the object is not deformed.
    \item \textbf{Low Resolution}: The texture appears as blurry in contrast with other elements of the scene at the same distance from the observer. This can occur when lower resolution texture is loaded with a different LoD (Level of Detail) e.g. In practice we recreate this glitch by scaling the texture uniformly in every direction (see Figure \ref{fig:Sample_LowRes}).
\end{itemize}

\textbf{Missing Texture}: One or more of the textures from the object are not rendered at all, which can be noticed in two ways. 

\begin{itemize}
    \item \textbf{Missing}: When the texture is not rendered due to a malfunction in the object the graphical engine renders no details like texture or shadows. This results that the space where the object is located appears as a solid color. This is an easy glitch to reproduce as removing the texture will trigger a standard function in respectively game engine. E.g. in Unity3D the replacing color is a pink monochromatic placeholder (see Figure \ref{fig:Sample_Missing}).
    \item \textbf{Placeholder}: Often the texture may not be available due to many different reasons. E.g. the texture was not correctly loaded, not found in the asset database, not transferred, etc. In these cases, while developing the game a default placeholder texture is used. These are not supposed to be used in the final version of the game but sometimes they are not updated to the correct ones resulting in faulty images (Compare Figure \ref{fig:Sample_PlaceholderWhite} and \ref{fig:Sample_PlaceholderPattern}). To generate these data samples, we substitute the textures in the object for a placeholder. In order to assess how the choice of texture affects the performance of the algorithm two placeholder versions were generated one plain white (corresponding to the default texture in the used game engine, Figure \ref{fig:Sample_PlaceholderWhite}) and another with a checked pattern and letters that indicate that is a default texture (Figure \ref{fig:Sample_PlaceholderPattern}). The latter solution is often used in order to be more visible to human testers.
\end{itemize}

\begin{figure}[t]
    \centering
    \begin{subfigure}[b]{0.23\textwidth}
        \centering
        \includegraphics[width=\textwidth]{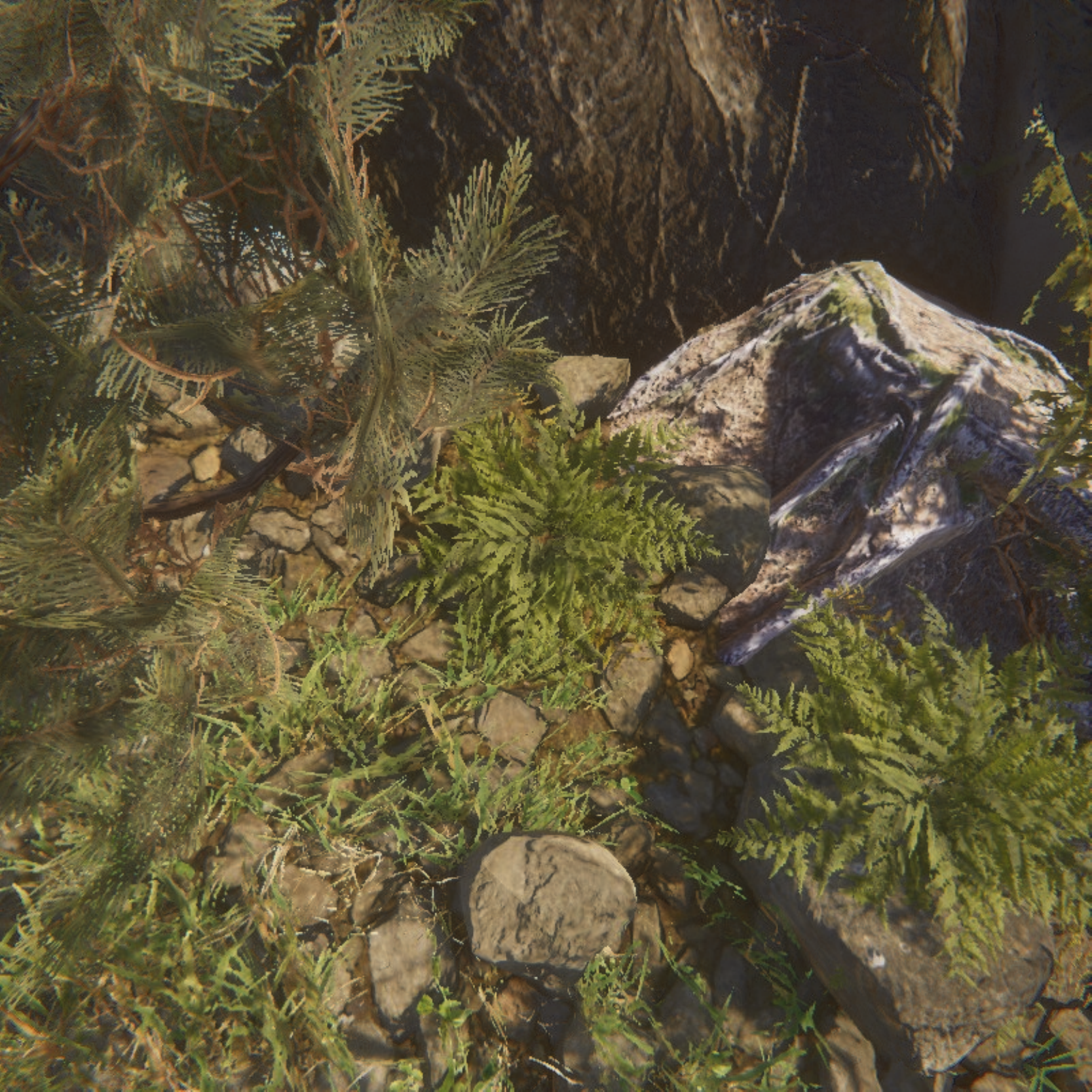}
    \end{subfigure}
    \hfill
    \begin{subfigure}[b]{0.23\textwidth}  
        \centering 
        \includegraphics[width=\textwidth]{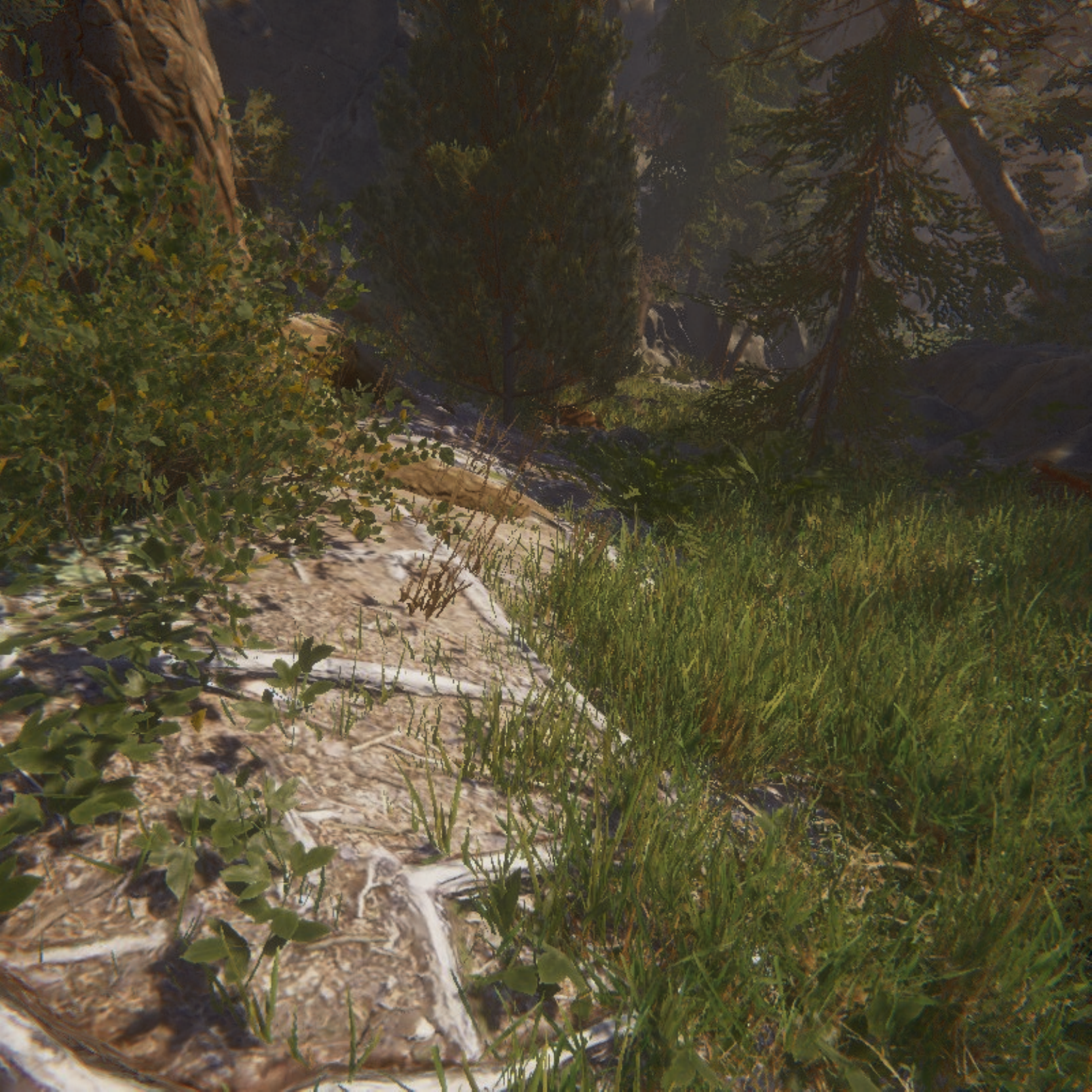}
    \end{subfigure}
    \caption{Additional normal samples from {\it Book of the Dead} Environment}
    \label{fig:Samples_BOTD}
\end{figure}

\begin{figure}[t]
    \centering
    \begin{subfigure}[b]{0.23\textwidth}
        \centering
        \includegraphics[width=\textwidth]{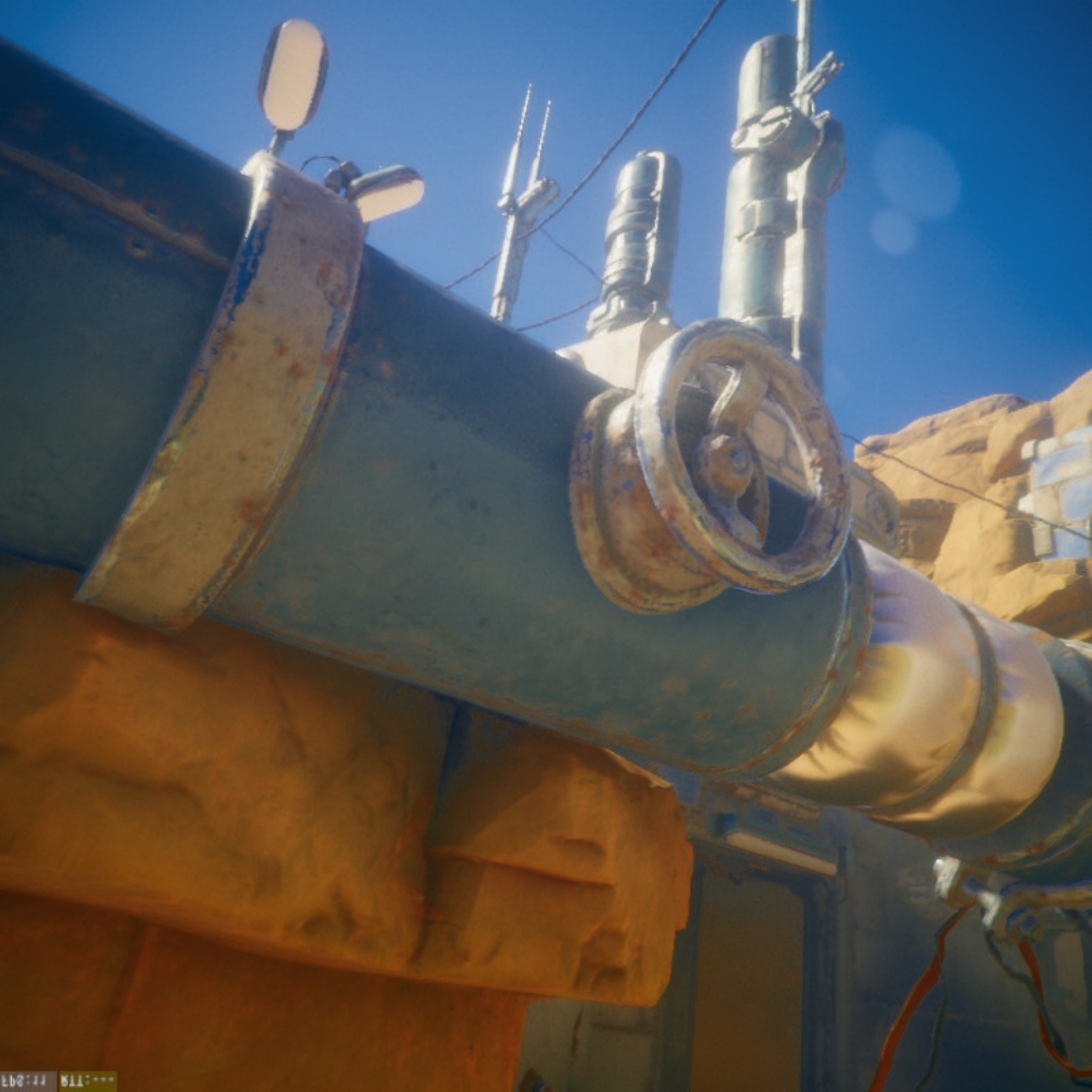}
    \end{subfigure}
    \hfill
    \begin{subfigure}[b]{0.23\textwidth}  
        \centering 
        \includegraphics[width=\textwidth]{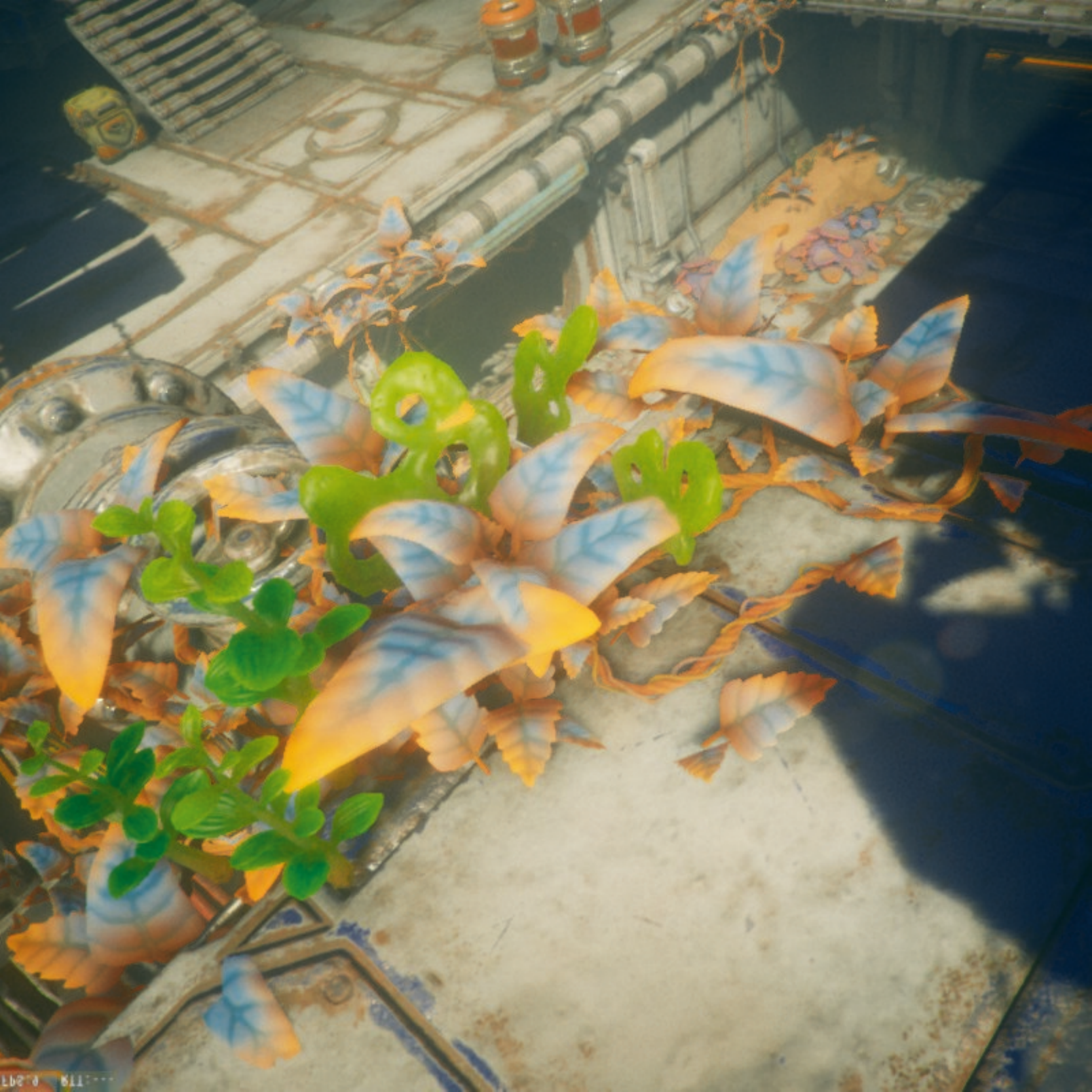}
    \end{subfigure}
    \vskip\baselineskip
    \begin{subfigure}[b]{0.23\textwidth}
        \centering
        \includegraphics[width=\textwidth]{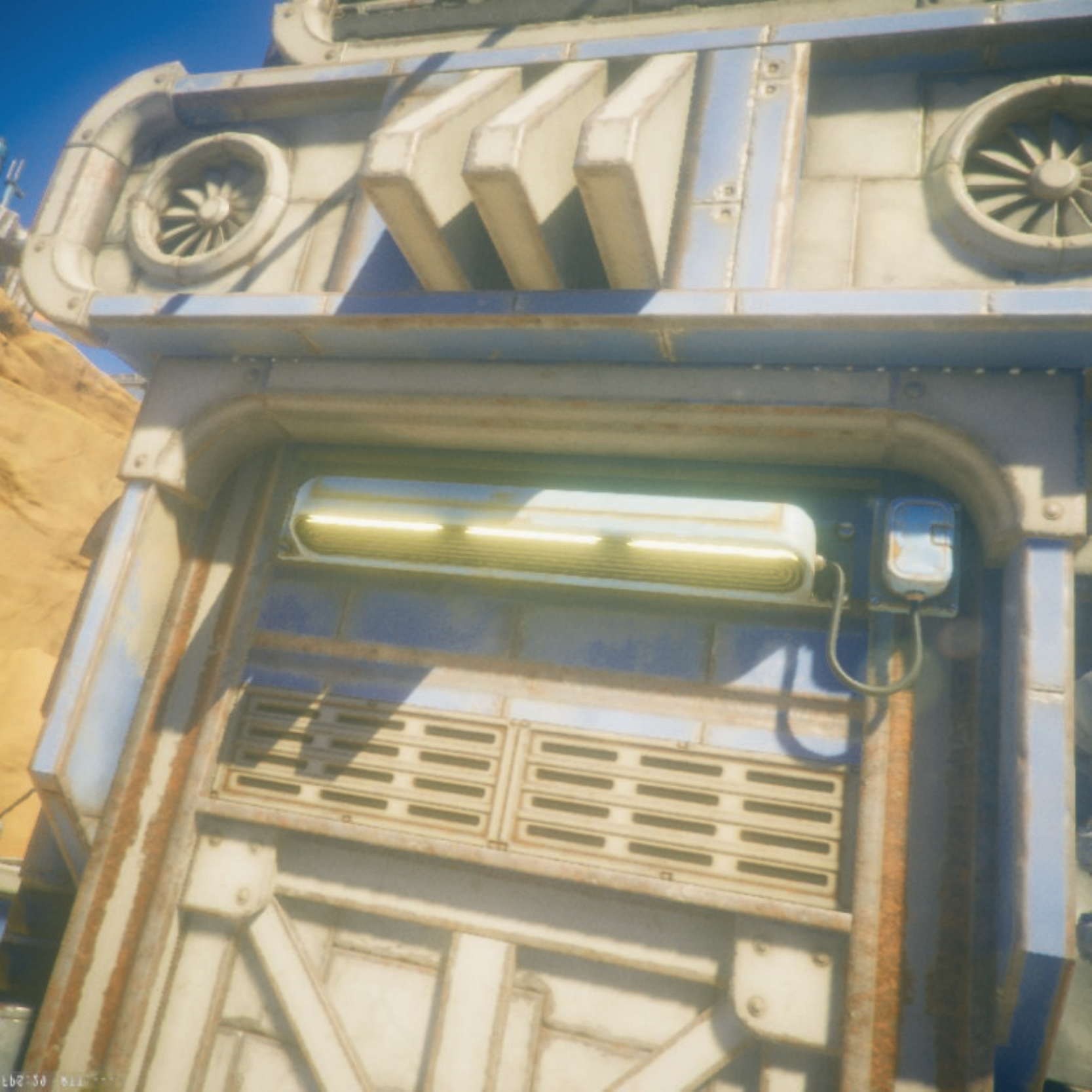}
    \end{subfigure}
    \hfill
    \begin{subfigure}[b]{0.23\textwidth}  
        \centering 
        \includegraphics[width=\textwidth]{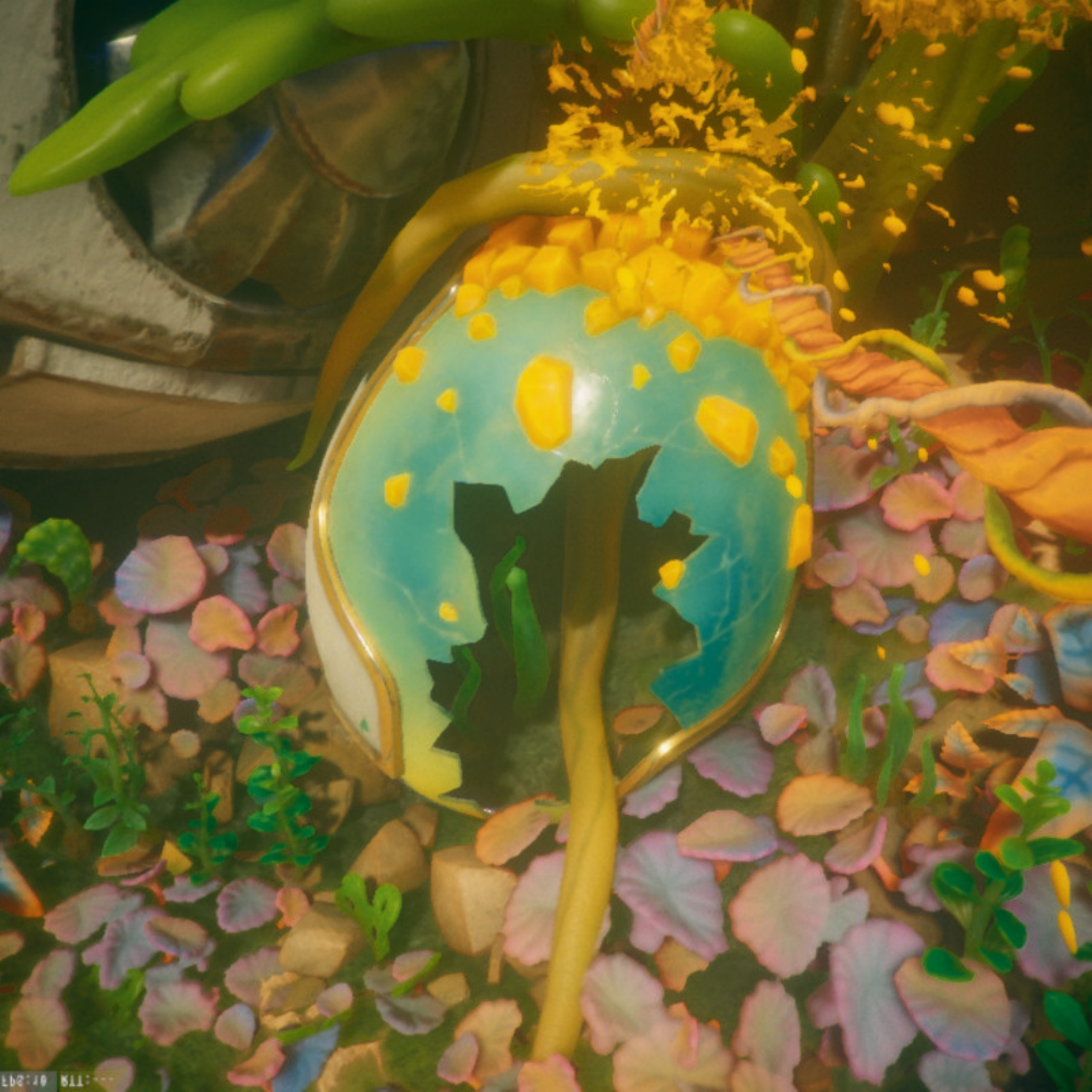}
    \end{subfigure}
    \caption{Normal samples from {\it FPS Sample} Environment}
    \label{fig:Samples_FPS}
\end{figure}

\begin{figure}[ht]
    \centering
    \begin{subfigure}[b]{0.23\textwidth}
        \centering
        \includegraphics[width=\textwidth]{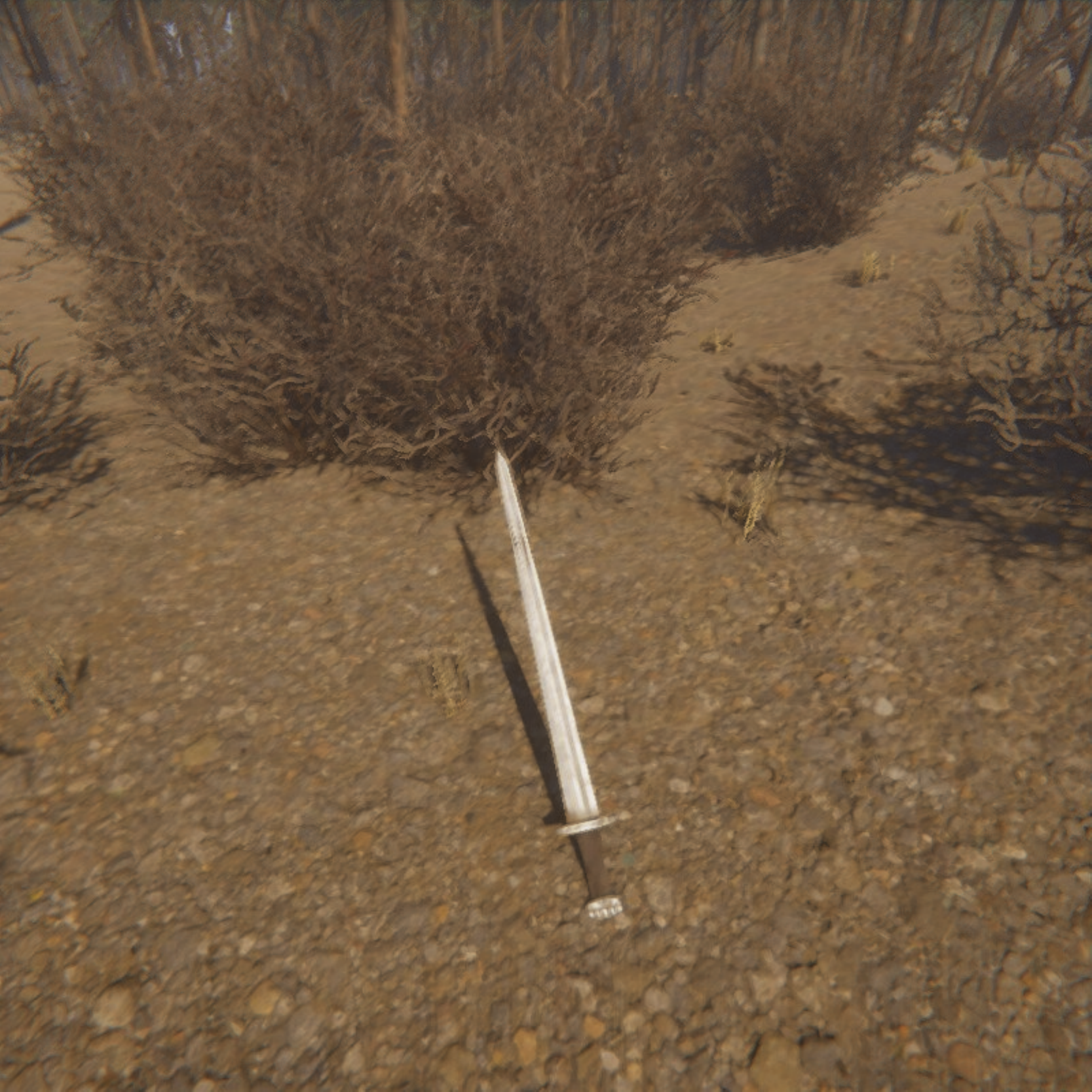}
    \end{subfigure}
    \hfill
    \begin{subfigure}[b]{0.23\textwidth}  
        \centering 
        \includegraphics[width=\textwidth]{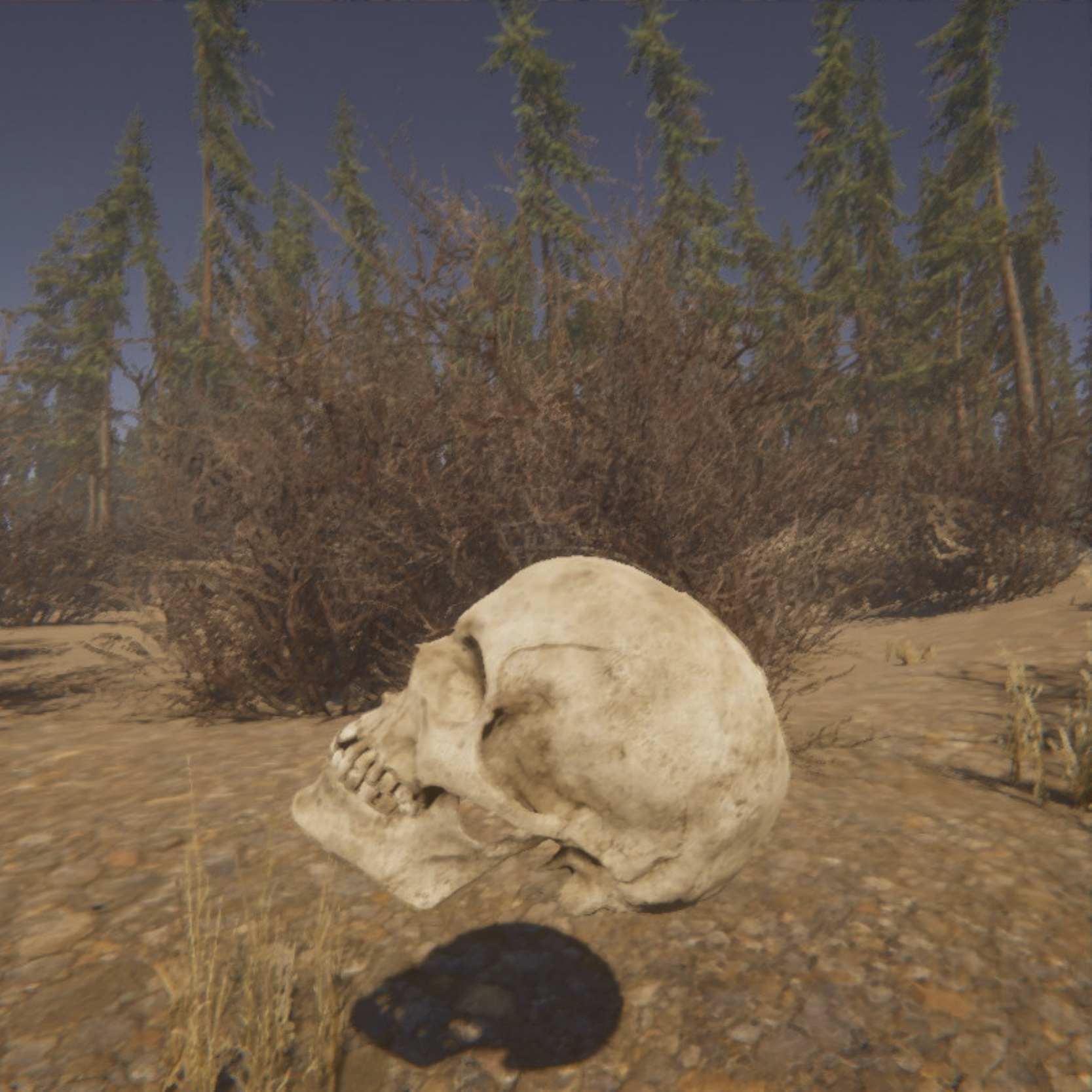}
    \end{subfigure}
    \vskip\baselineskip
    \begin{subfigure}[b]{0.23\textwidth}
        \centering
        \includegraphics[width=\textwidth]{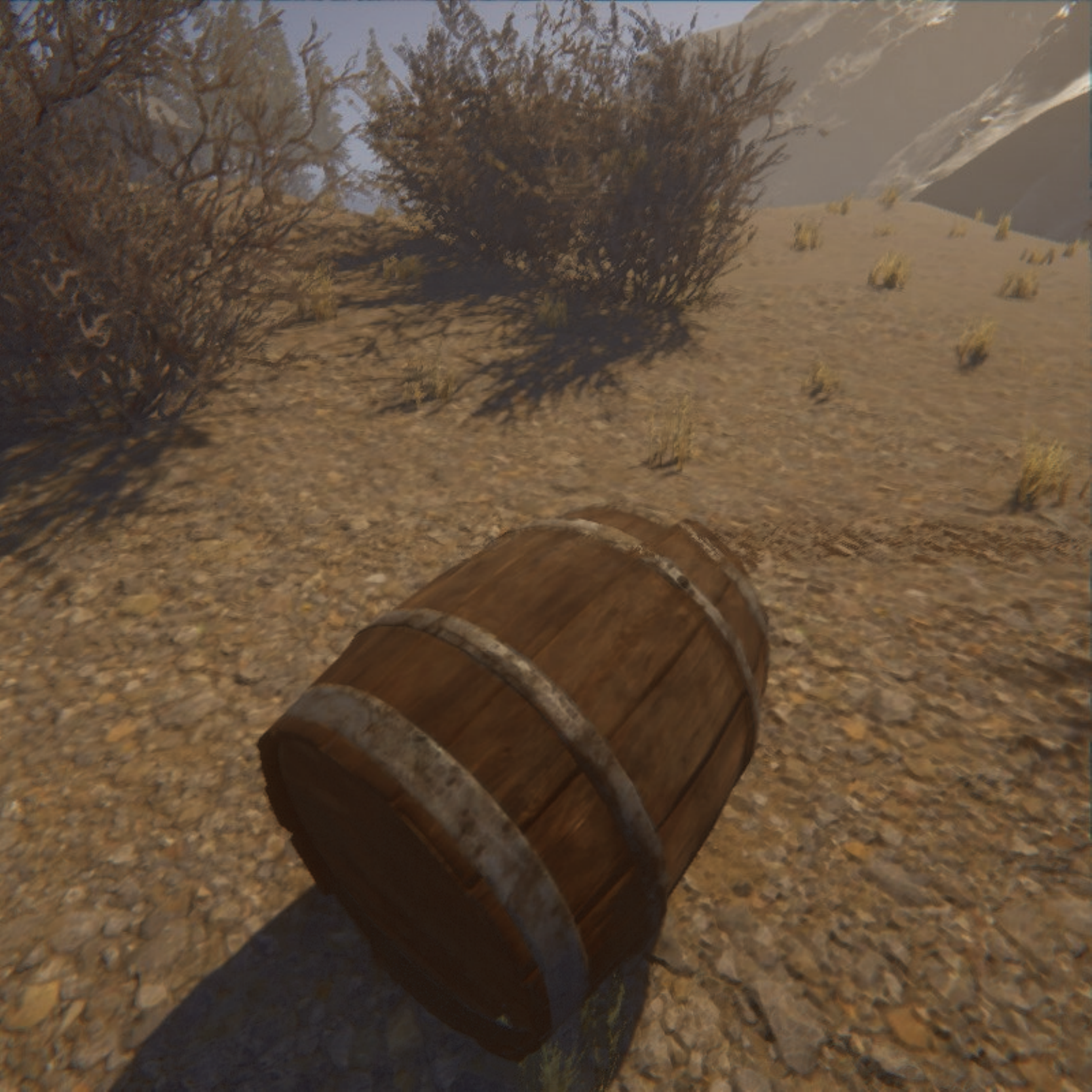}
    \end{subfigure}
    \hfill
    \begin{subfigure}[b]{0.23\textwidth}  
        \centering 
        \includegraphics[width=\textwidth]{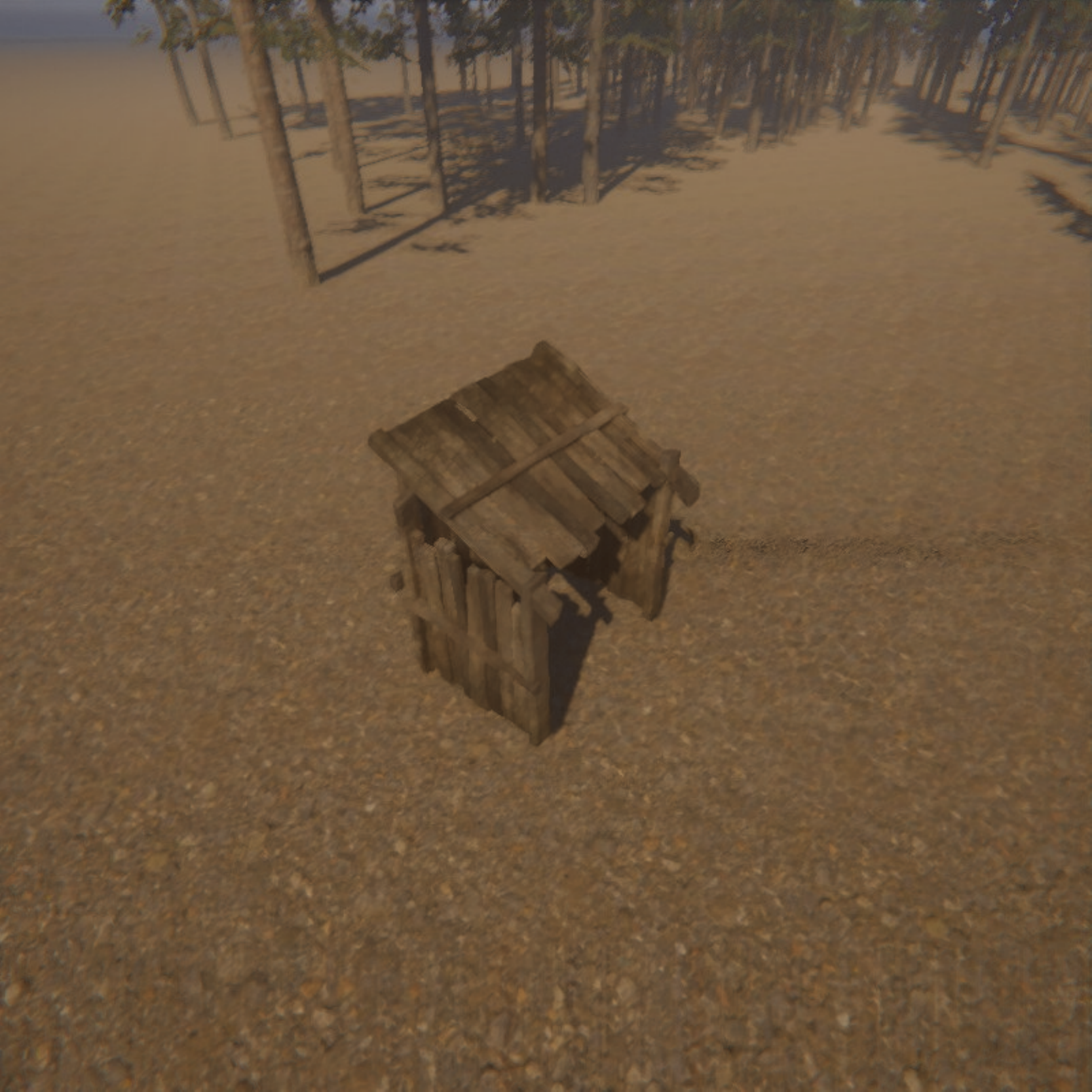}
    \end{subfigure}
    \caption{Normal samples from {\it Viking Village} Environment}
    \label{fig:Samples_VikingVillage}
\end{figure}

The dataset generated was balanced with a size of 12700 samples, 20\% of the data were normal samples (Figure \ref{fig:Samples_BOTD}) and 80\%  different kinds of glitches. To capture all the images 127 objects were used meaning that 100 images were taken per object in various angles and distances. The data was generated using assets from the {\it Book of the Dead} available in the Unity3D store. Also, two extra data sets were obtained in the same fashion with elements from Unity3D {\it FPS Sample} (Figure \ref{fig:Samples_FPS}) and {\it Viking Village} (Figure \ref{fig:Samples_VikingVillage}) also available in the Unity Asset Store, creating two more data sets with respectively 10000 and 11500 samples. In total 34200 samples were created. These extra data sets were generated to assess the generalizing ability of the method described.

\section{Methodology}

The method proposed uses a DCNN classifier to determine whether an image contains a glitch or not. As an input an 800x800 color image is provided to the network, this resolution was chosen balancing efficiency, and the use case. The image is forward passed through a CNN and a vector with 5 components is obtained as an output, these represent the probability of the image being normal or a glitch, dividing the glitch class in 4 attending on the glitch type

\subsection{DCNN Architectures}
The network structure for the classifier is one of the most important aspects to consider. To base our decision, we looked at the state-of-the-art networks in classification, paying attention to the performance in ImageNet. We argue that this data set is relevant to our problem since images in video games mimic to a large extent the real world. It is this property that motivates why these methodologies also perform well in the formulated problem. 

While ImageNet accuracy was an important factor, we also considered the computational requirements of the network. Avoiding networks with excessive computation time will allow running the model in parallel with other tasks, like the rendering of the video game images. We trained state-of-the-art architectures such as VGG, Alexnet, ResNet etc. 
Based on this we focused on two representative architectures ResNet \cite{he2016deep} and ShuffleNetV2 \cite{ma2018shufflenet} which showed most promising considering accuracy and computational resources required. For ResNet different degrees of complexity were explored by training ResNet18, 34 and 50 with the number indicating the number of convolutional layers. Similarly, for ShuffleNetV2 we explored the versions x0.5, x1.0, x1.5, x2.0, in increasing order for the number of filters in the layers.

\subsection{Training}
The implementation of the models was done in Pytorch, based on the implemented models in the Torchvision library. The models were modified to adjust to the number of classes in our problem. In order to fit the model to our data the Adam optimizer \cite{kingma2014adam} was used, minimizing the cross-entropy loss function, commonly used in the classification problem. The hyper-parameters of the optimizer such as the batch size and the learning rate were explored in order to determine which was the best configuration for this problem. The models were trained for 100 epochs, reaching stable performance after 50 epochs or less. 

Due to the similarity with real images and the known power of pre-trained models \cite{huh2016makes} the initialization of the networks was done in two different ways: random initialization, using the standard initialization, and pre-trained weights on ImageNet as starting point.

The core data used was provided by the environment from the {\it Book of the Dead} assets, however, in order to asses the capacity of the model additional data from the other environments (mentioned in the previous section) was also used. Also, the model was applied to other data sets to asses how the approach would work in different environments.

To validate the model the separation between training and validation data set was done attending to the objects. This means that all the images correspondent to one object are present in either the training or validation set but not in both at the same time, which gives more validity to the performance observed in the validation set. For all the training runs 20 \% of the objects are placed in the validation set while the remaining 80\% is used for training. Although different splits were used obtaining similar results, in the results section, the same partition is used allowing for direct comparison between the different training configurations. 

\subsection{Metrics}
To assess the performance of the model, different metrics were deployed. These metrics should be intuitive and easy to interpret giving allowing for a correct understanding of the model is performing \cite{DavidW2011Evaluation}. The cross-entropy loss was used to fit the networks, but its interpretation is not straightforward since it takes arbitrarily high values, and it does not only depend on whether the prediction obtained matches the original label but also on the probability assigned to this prediction. Instead, we will take accuracy as the main metric to compare different models. Since the data we are using is balanced by construction (all the classes exist in the same proportion) both in the training and validation set, accuracy is not misleading regarding the performance of the model. The accuracy is simply computed as the fraction of samples that whose classes were correctly predicted.

Since we can consider this a binary problem (i.e. normal image/glitch image), we can define metrics like: {\it precision} (fraction of the elements predicted as glitches that are actually glitches), {\it recall} (fraction of the actual glitches that are predicted as glitches), and {\it false positive rate} (fraction of the negative samples that are predicted to be positive). These metrics will be provided for the best performing model. 

Confusion matrices provide detailed information about the behavior of the network. They group the elements into a matrix attending to the original class (rows) and the predicted class (columns). This matrix will be normalized with respect to the original class, meaning that the elements in a row will always add to one. The ideal classifier would have a confusion matrix with ones at the diagonal elements. The confusion matrix provides visibility into which are the classes that are mistaken, giving an insight into why the network may be failing.

\begin{table*}[ht]
\selectfont
\centering
\caption{Comparison of the performance and computation time for the different networks considered. All networks were trained using batch size (8) and learning rate ($10^{-3}$)}\label{tab:Results_Network}

\begin{tabular}{lllll}
\hline
\textbf{Network}                            & ResNet18      & ResNet34      & ResNet50      & ShuffleNetV2  \\ \hline
\textbf{Accuracy (Random Init. Validation)} & 0.674 (0.012) & 0.642 (0.006) & 0.671 (0.018) & 0.617 (0.007) \\
\textbf{Accuracy (Pretrained Validation)}   & 0.721 (0.014) & 0.719 (0.017) & 0.724 (0.031) & 0.767 (0.014) \\
\textbf{Accuracy (Pretrained Training)}     & 0.958 (0.006) & 0.951 (0.007) & 0.886 (0.015) & 0.970 (0.004) \\
\textbf{Execution Rate}                     & 60 fps        & 60 fps        & 45 fps        & 60 fps        \\ \hline
\end{tabular}
\end{table*}

\begin{table*}[ht]
\selectfont
\centering
\caption{Exploration of the learning rate parameter: ShuffleNet v2 with constant batch size 8}\label{tab:Results_LR}
\begin{tabular}{lllll}
\hline
\textbf{Learning Rate}         & 10-2          & 10-3          & 10-4          & 10-5          \\ \hline
\textbf{Accuracy (Validation)} & 0.715 (0.015) & 0.767 (0.014) & 0.801 (0.012) & 0.791 (0.007) \\ \hline
\end{tabular}
\end{table*}

\section{Results}
\label{sec:results}

We present the results from training the DCNNs with the setting stated in the previous section. Note that the results presented in the tables present two numbers. The first correspond to the average of the metric during the last 20 epochs and second is standard deviation (in parenthesis). Similarly, the confusion matrices present the average of the 20 last epochs, providing a more stable result. 

First, we compared different DCNN architectures and decided to focus in two networks that provided the best results while being most performant: ResNet and ShuffleNet v2. 
Table \ref{tab:Results_Network} presents a comparison between these selected models; here, we focus on the accuracy and we highlight the differences between using the weights of a network pre-trained in ImageNet as a starting point and a random initialization. We also show the differences for this metric between the training set and the validation set. Another feature which is relevant is the execution time.

\begin{table*}[t]
\centering
\caption{Exploration of the batch size parameter: ShuffleNet v2 with constant learning rate $10^{-4}$ }\label{tab:Results_BS}
\begin{tabular}{lllll}
\hline
\textbf{Batch Size}            & 4               & 8               & 16              & 32              \\ \hline
\textbf{Accuracy (Validation)} & 0.791   (0.012) & 0.804   (0.012) & 0.805   (0.011) & 0.819   (0.006) \\ \hline
\end{tabular}
\end{table*}

\begin{table*}[t]
\centering
\caption{Performance on different data sets: ShuffleNet v2, batch size 32 and learning rate $10^{-4}$  }\label{tab:Results_DD}
\begin{tabular}{lllll}
\hline
\textbf{Dataset}               & Book of the Dead & FPS Sample    & Viking Village & All (Val on BotD) \\ \hline
\textbf{Accuracy (Validation)} & 0.819 (0.006)    & 0.852 (0.010) & 0.807 (0.005)  & 0.802 (0.008)     \\ \hline
\end{tabular}
\end{table*}

\begin{figure}[ht]
    \centering
    \includegraphics[width=0.45\textwidth]{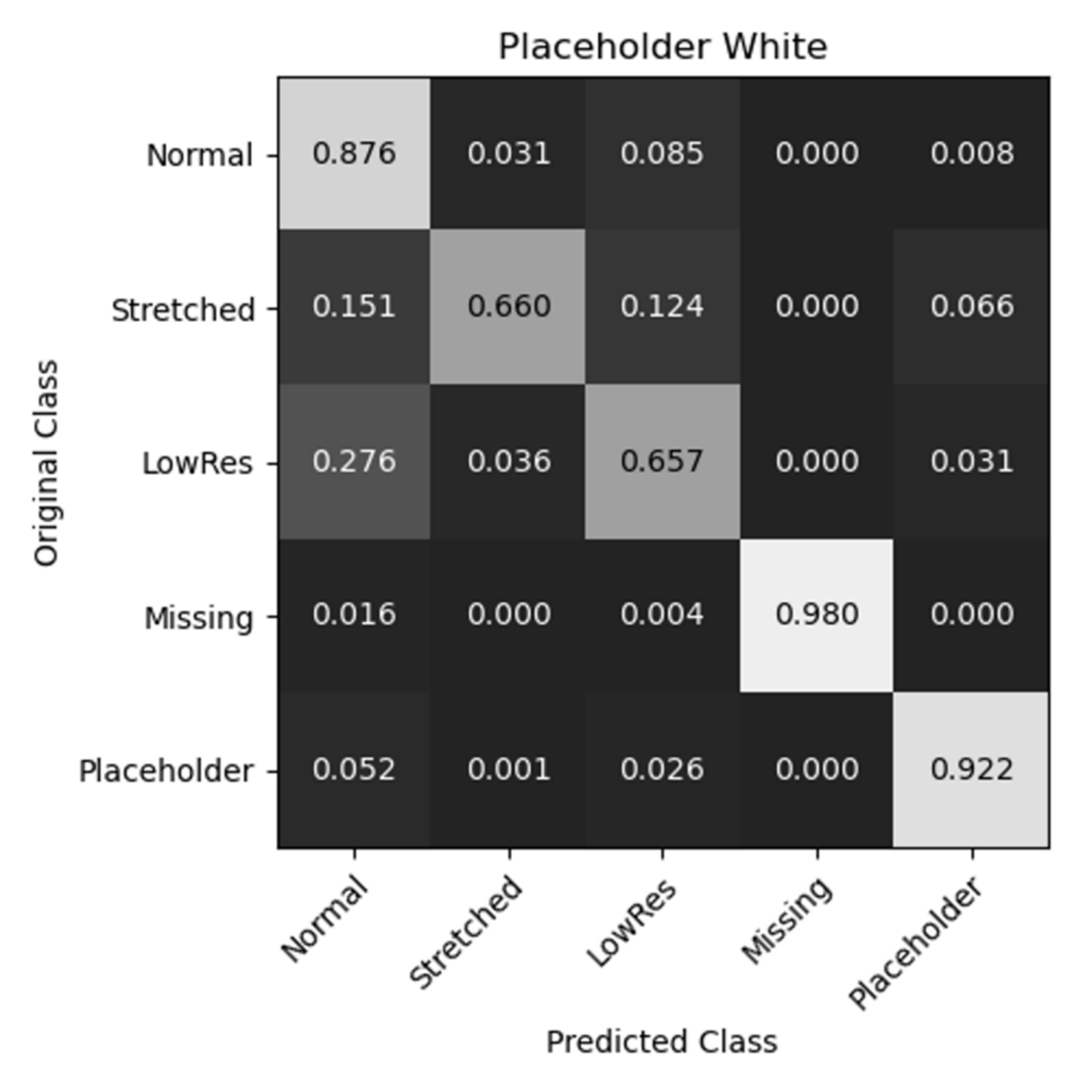}
    \caption{Confusion Matrix: White Placeholder}
    \label{fig:Results_CM_PlaceholderWhite}
\end{figure}

The results show some significant findings. First, we see an improvement when using pretrained models for all the networks, which can be linked to the similarity of our samples to natural images. Another characteristic noted is the difference between the performance in training and validation, we can clearly see that the model overfits the data. This indicates that the models may be too big in relation to the amount of data, which leads to overfitting. Note that the data set used in this paper was relatively small with 12700 samples. 
Lastly, we found that the model which works best for this task is ShuffleNetV2 due to both, its better performance and execution time. As a note, we would like to state that complexity regarding the ShuffleNetV2 structure was assessed and as with ResNet no significant differences in performance were discovered. Other architectures were also evaluated: VGG \cite{simonyan2014very} and AlexNet \cite{krizhevsky2012imagenet} provided worse results, while MobileNet v2 \cite{sandler2018mobilenetv2} achieved similar performance as ShuffleNet v2, although with slower training times. 
We would like to remark that this is an open problem in computer vision and incoming research may provide more accurate and fast architectures.

We also explored the hyperparameter space when fitting the models with a focus on the network that we found to perform best, i.e. ShuffleNetV2. To explore the learning rate, we tried 4 different values equally spaced in a logarithmic scale. The results are compiled in Table \ref{tab:Results_LR} focusing in the accuracy of the validation set. There we see that the optimal value for this task was $10^{-4}$.

The batch size used in the Adam optimizer was also explored, again with equally spaced values in a logarithmic scale. The results are presented in Table \ref{tab:Results_BS}. The maximum batch size used was restricted by the graphics card used but we can see the pattern of an increasing accuracy with the batch size.

\begin{figure}[ht]
    \centering
    \includegraphics[width=0.45\textwidth]{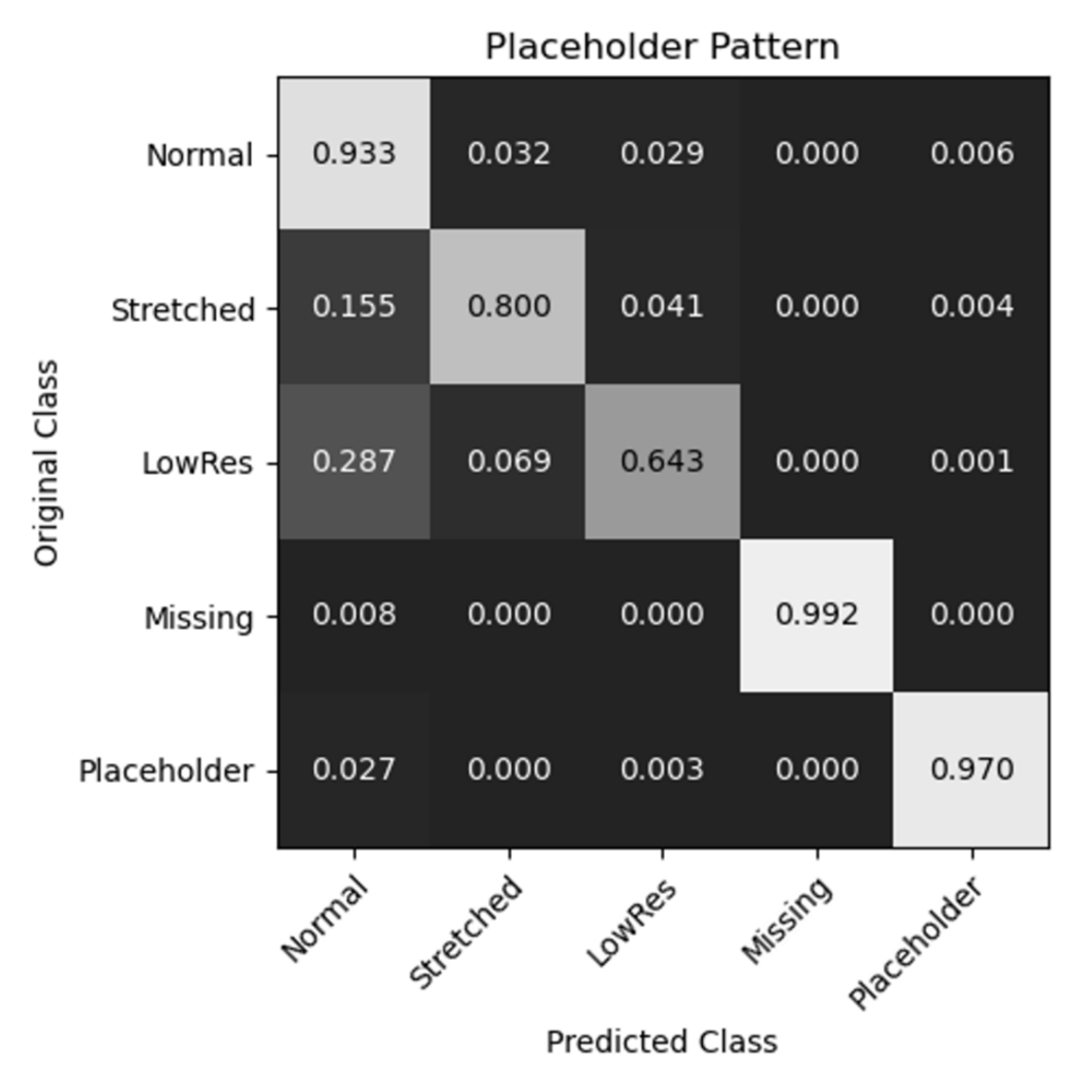}
    \caption{Confusion Matrix: Pattern Placeholder}
    \label{fig:Results_CM_PlaceholderPattern}
\end{figure}

Another aspect that improved notably the performance of the networks was using better data, the clearer improvement was regarding the placeholder texture. When using a white placeholder texture, the network learns to recognize white elements as glitches, misclassifying images with white elements (Figure \ref{fig:Results_CM_PlaceholderWhite}). A bit surprisingly, using a recognizable pattern for the placeholder reduced misclassification in other classes as well, this can be seen in Figure \ref{fig:Results_CM_PlaceholderPattern}. We reason this is due to the fact that low-res and stretched textures can sometimes resemble a plain texture allowing the network to classify low-res and stretched texture better when it does not resemble the placeholder one. In general the confusion matrices allow us to see which classes the network struggles to classify. The glitches with the best performance correspond to those that are easily perceived the missing texture group, while the corrupted textures present more confusion.

The best performance achieved was when using a recognizable pattern as a placeholder texture (Figure \ref{fig:Results_CM_PlaceholderPattern}). Here we achieve an average (for the last 20 epochs) of 86.8\% accuracy with a 6.7\% false positive rate and a recall of 88.1\%. Note that for the recall we consider the binary problem, there is no difference if a glitch is predicted in the wrong glitch class if it is predicted as a glitch. 

Next we present the results for the other environments. In the Data section we describe how we generated other data sets, in Table \ref{tab:Results_DD} we can see the performance of the proposed framework with different data sets, seeing how this is similar for all the data sets, this indicated that this method may be successfully applied to different video game styles. With {\it Book of the Dead} and {\it Viking Village} presenting more realistic looking assets, while the dataset based on {\it FPS Sample} represent a stylized looking game. Although, the styles of video games represented in this study are not extensive and further research may be needed to assess if this can be generalized to other types of environments. 
We also provide the performance on {\it Book of the Dead} when the data set is trained in all the data sets here we see that the performance is lower, which indicates that is advisable to train the model in the environment that it will be applied rather than having one general model. 

\subsection{Refined model}

In this subsection we present modifications applied in our work and briefly discuss their potential contribution in the problem formulated.

\subsubsection{Grouping}
The glitches generated can be divided into two groups: corrupted and missing textures. The problem is then defined as a 3 class classification problem. Also, considering only correct and faulty types, a binary classification problem can be defined. These other setting were tried with no improvement in performance. As the differences in computation and memory costs are negligible between these formulations, the 5 class setting was deemed optimal, since it provided more information.

\subsubsection{Semantic Segmentation}

The classification approach provides limited information on the glitches detected, since no localization is provided. In contrast, Semantic Segmentation can provide pixel-wise classification and has been object of successful research in recent years \cite{garcia2017review}. Thus, we have also considering relevant state of the art Segmentation structures to approach this problem: U-Net \cite{ronneberger2015u}, Fully Convolutional Networks (FCN) \cite{long2015fully} and Deep Lab V3 \cite{chen2017rethinking}.

Although highly informative, this approach is computationally and memory expensive, and gave substantially worse results in this problem domain. Obtaining values for the Detection Rate of 0.266, 0.441, and 0.559 respectively for the U-Net, FCN and Deep Lab V3 structures, defined as the fraction of the glitches with a Intersection over Union score higher than 0.5, much lower when comparing with classification and taking Recall as an equivalent metric.

Although the low performance could be partially explained due to difficulties found in defining an adequate label, this setting was deemed not the most adequate in comparison with the classification approach.

\subsubsection{Aggregate images}
\label{sec:aggregated_images}

One of the biggest advantages when working with video game images and anomaly detection is the total control over the camera. This provides our models with more information for each object, like capturing several images from different angles, lighting conditions, distances, etc. In Table \ref{tab:Results_Agreggating Images} we present the results obtained when using several images for classifying an object. The probabilities for each class are computed averaging the probabilities obtained from the network. We do notice an improvement even when just using 2 images instead of 1, note that this slows down as the number of images increases.

\begin{table}[t]
\centering
\caption{Performance aggregating images from the same object for classification. }\label{tab:Results_Agreggating Images}
\begin{tabular}{lllll}
\hline
\textbf{\# Images} & 1     & 2     & 10    & 20    \\ \hline
\textbf{Accuracy}  & 0.832 & 0.851 & 0.861 & 0.864 \\ \hline
\end{tabular}
\end{table}

\paragraph{Confidence Measures}

In a real use case when testing a video game, particularly in later phases of the development, glitches may be scarce, which makes preventing false positives important. One of the problems of Neural Networks in classification is that the probability for each class has to add to one, providing high probabilities even when doing big extrapolation. 

In this work we have investigated the confidence measure presented in \cite{hess2020softmax} (Gauss-Confidence) for its ability to reduce false positives. When applied to this problem we obtain higher confidences for for missing and placeholder glitches having mean values of 0.643 and 0.633  respectively, and lower values of 0.259 and 0.202 for stretched and lower resolution textures respectively. From this we propose using the confidence measure as a potentially effective method to filter out false positives.

Furthermore, when using this the confidence measure could be used as a signal to acquire more images (using the same technique in previous section) from the object in order to increase the confidence.

\section{Future Work}

As mentioned, one of the shortcomings of this approach is the lack of localization information. Although semantic segmentation was tested further research may be conveyed to further assess the viability of this approach. Object detection could be a good alternative to provide certain localization information, together with fast computation time, as an area with broad successful research in recent years  \cite{zou2019object} and structures like YOLO V4 \cite{bochkovskiy2020yolov4} that allow for real time object detection.

 The presented approach mainly recognizes the type of glitches which the network was trained on, thus it will tend to miss issues which was not in the training data. Here unsupervised and semi-supervised anomaly detection approaches could offer a way of detecting unseen graphical anomalies. Nevertheless, applying these methods is not a trivial task especially when dealing with high dimensional data as it is in the case of images but it is certainly an interesting venue of research. 
 
 Furthermore, combining this graphical ML approach with logical testing using reinforcement learning {\it \cite{bergdahl2020augmenting}} could offer interesting synergy effects. For example, connecting the reward signal of an automated testing agent to the number of graphical glitches it finds, can potentially make it more effective in seeking out graphical bugs in the game at the same time providing more training data to the supervised model.

\section{Conclusion}

This study presents an approach for detecting graphical anomalies in video game images. Making use of a supervised approach with a DCNN we were able to detect 88.1 \% of the glitches with a false positive rate of 6.3 \%. These results give confidence that we can use such models in a testing environment to detect graphical malfunctions. On top of that several considerations regarding data, architecture, hyper-parameters, and network initialization are discussed providing general guidelines for future methods addressing the problem. We have seen that this method is highly data-dependent with better generalization with a growing amount of data. On the other hand in a modern game with thousands of assets and environments, generating a diverse training set is relatively easy which makes this a valid approach in modern game production.

Therefore, the conclusion of this paper is that DCNNs can be used in a production environment to improve the testing process in video games, partially automating a now manual process. The architecture used was able to evaluate images at a rate of 60 fps, being computationally light enough to run in parallel with the rendering allowing for real-time assessment on the video quality. Another interesting find is that default fall-back placeholder textures with a recognisable pattern to a human, is also improving the classification rate significantly. Unlike blob detection techniques that detects certain colors, the same reasoning applies as for humans: in order to more easily detect errors, it is better to use a recognizable pattern than a plain color. Also, we found that low resolution textures scored the lowest accuracy is something that human testers also tend to get wrong. Similarly, from experience, human testers tend to classify low resolution as false-positive more than the other classes.

Furthermore, since the methods presented are trained directly in the images from the game, glitch detection is presented as an independent tool, providing high versatility when used in the actual testing pipeline. This will allow for diverse uses like processing all the images during a game play, scripting a camera focusing only on specific instances, or running completely separated from the game using recorded images.

One of the main limitations of the method presented is being supervised which means that specific data exemplifying both normal and anomalous images must be provided. Although some level of extrapolation was displayed by the model, this approach would be less useful during the first stages of game development since no data is available. These methods become more effective in the later stages of the development when operations like migrating to a new version (other console platforms / graphical engine updates) or adding new objects to the game, both cases in which sample data is available to train the model.

On the other hand, in this work, we have only explored one of many possible approaches to the problem, by restricting ourselves to a very particular subset of glitches. Further research is needed to asses whether other problems in testing could also be addressed, focusing on other kinds of glitches, and not only limited to static images but also including video sequences. 

\section{Acknowledgements}

The authors would like to thank Cristian Deri, Jan Schmid, Jesper Klittmark (EA DICE) for the valuable feedback and providing us with data. 

\bibliography{refs}
\bibliographystyle{aaai}

\end{document}